\pdfoutput=1
\documentclass[10pt,logo,copyright]{nvidia_release_template/paper_nvidia_report}
\usepackage[authoryear,sort&compress,round]{natbib}
\setcitestyle{numbers}
\usepackage[font=footnotesize]{caption}

\correspondingauthor={}

\usepackage{amsmath,amssymb}
\usepackage{array}
\usepackage{booktabs}
\usepackage{colortbl}
\usepackage{enumitem}
\usepackage{algorithm}
\usepackage[noend]{algpseudocode}
\usepackage{etoolbox}
\makeatletter
\patchcmd{\ALG@doentity}{\item[]\nointerlineskip}
  {\item[]\vskip-\prevdepth\nointerlineskip}{}
  {\PackageWarningNoLine{main}{Could not patch algorithmicx's noend spacing}}
\makeatother
\usepackage{xspace}
\usepackage{cleveref}
\usepackage{tcolorbox}
\tcbuselibrary{skins}
\DeclareFontFamily{T1}{JetBrainsMono}{}
\DeclareFontShape{T1}{JetBrainsMono}{m}{n}{<-> nvidia_release_template/JetBrainsMono-Font-TTF/JetBrainsMono_Rg}{}
\pdfmapline{+JetBrainsMono_Rg < nvidia_release_template/JetBrainsMono-Font-TTF/JetBrainsMono-Regular.ttf <T1-WGL4.enc}

\newcommand{\paraspacebefore}{-0.5em}
\newcommand{\paraspaceafter}{-0.5em}
\newcommand{\capspacebefore}{-0.5em}
\newcommand{\capspaceafter}{-0.5em}
\newcommand{\tabspacebelow}{-0.5em}

\crefname{equation}{Eq.}{Eqs.}
\crefname{figure}{Fig.}{Figs.}
\crefname{section}{Sec.}{Secs.}
\crefname{appendix}{App.}{Apps.}
\crefname{table}{Tab.}{Tabs.}
\crefname{algorithm}{Alg.}{Algs.}

\definecolor{cvprblue}{rgb}{0.21,0.49,0.74}
\definecolor{refred}{rgb}{0.70,0.18,0.16}
\hypersetup{colorlinks=true,allcolors=cvprblue,linkcolor=refred}
\definecolor{tablefamily}{HTML}{F1F2EF}
\definecolor{tableours}{HTML}{E8F3D6}
\newcolumntype{Y}{>{\centering\arraybackslash}X}

\DeclareRobustCommand{\method}{Recova\xspace}
\definecolor{avocadoGreen}{HTML}{568203}

\DeclareFontFamily{T1}{RecovaSans}{}
\DeclareFontShape{T1}{RecovaSans}{m}{n}{<-> nvidia_release_template/NVIDIA-Sans-Font-TTF/NVIDIASans_Md}{}
\DeclareRobustCommand{\methodtitle}{%
  \recovaicon\scalebox{1.12}{\fontencoding{T1}\fontfamily{RecovaSans}\fontseries{m}\fontshape{n}\selectfont\color{avocadoGreen}Recova}}
\title{\methodtitle{}: Agent-Guided Failure Recovery\\ for Autonomous Robotic Manipulation}
\author{
\parbox{\linewidth}{
\raggedright
{\small\bfseries
Isabella Liu$^{1}$,
An-Chieh Cheng$^{1}$,
Johan Bjorck$^{3}$,
Zhiding Yu$^{3}$,
Hongxu Yin$^{3}$,}\\
{\small\bfseries
Jan Kautz$^{3}$,
Linxi Fan$^{3}$,
Yuke Zhu$^{2,3}$,
Sifei Liu$^{3}$}\\
{\footnotesize
$^{1}$University of California, San Diego \quad
$^{2}$University of Texas at Austin \quad
$^{3}$NVIDIA}
}
}

\begin{document}

\maketitle
\addtocontents{toc}{\protect\setcounter{tocdepth}{0}}

\begin{center}
\begin{minipage}{\linewidth}
    \centering
    \includegraphics[width=\linewidth]{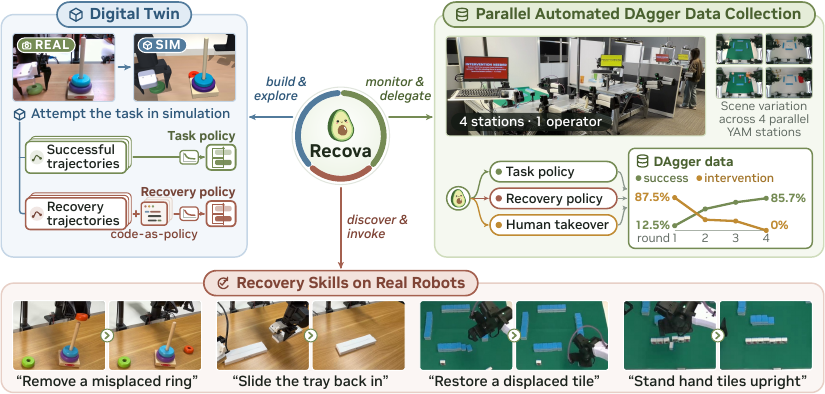}
    \vspace{\capspacebefore}
    \captionof{figure}{%
        \textbf{Overview of \method{}.}
        \emph{Digital twin (top left):} A coding agent explores task execution
        in a reconstructed workstation and develops recovery skills from failures.
        Task and recovery trajectories train separate policies, while recovery
        programs form a reusable skill library.
        \emph{Parallel DAgger (top right):} Across four real-robot stations, the
        agent coordinates execution and verified recovery, requests human help
        when needed, and uses the resulting experience to improve both policies.
        Success rises and human intervention falls over training rounds
        (\cref{fig:collection}).
        \emph{Recovery skills (bottom):} Paired frames show learned recoveries
        executed on real robots. Project page: \textbf{\url{https://www.liuisabella.com/Recova}}
    }
    \label{fig:teaser}
    \vspace{\capspaceafter}
\end{minipage}
\end{center}

\begin{abstract}
Manipulation failures can leave scenes in states from which a task policy
cannot recover. Learning corrective behaviors requires scalable failure
exploration and physical grounding. We present \method, an agent-guided
framework that jointly develops task execution and recovery in a reconstructed
digital twin, then verifies and refines both through real-world experience.
In the twin, the agent diagnoses failures, tests corrective programs, and
collects successful task and recovery rollouts for separate policies. During
deployment, it monitors progress, invokes a learned or programmatic recovery,
verifies scene restoration, and resumes execution. When no suitable recovery
is available, a human demonstration resolves the failure and enters the
learning loop, allowing the system to expand its recovery capabilities.
Physical rollouts and human demonstrations are routed to the corresponding
policy for DAgger training. Across six LIBERO-Pro settings and four
MolmoSpaces categories, \method achieves 78.8\% and 64.9\% mean success,
compared with 71.7\% and 38.0\% for the strongest baselines. With parallel
collection across four real-robot workstations, DAgger fine-tuning raises mean
success from 23.8\% to 77.5\%, and recovery skills further raise it to
87.5\%. Over four collection rounds on one task, observed human intervention falls
from 87.5\% to 0\%. Together, these results show how agent-guided recovery
turns failures into reusable capabilities, improving robustness while
progressively reducing human intervention.

\end{abstract}
\abscontent

\section{Introduction}
\label{sec:intro}

Vision-language-action (VLA) models enable robots to perform diverse
manipulation tasks from language instructions
\citep{brohan2023rt2,kim2024openvla,nvidia2025gr00t}, from folding laundry
and assembling boxes to cleaning unfamiliar homes
\citep{black2024pi0,pi2025pi05}. However, even a small mistake can halt an
otherwise capable policy. A tipped cup must be set upright before pouring can
continue; a pencil caught across a box rim must be moved before the lid can
close. In both cases, the robot must restore a workable scene before resuming
the task.

Most VLA training data, however, contain successful trajectories from
standard initial states and provide little guidance for restoring a failed
scene. Failures create a long tail of unusual configurations, and correcting
them may require undoing a previous action, repositioning an object, or
temporarily moving away from the nominal task goal. Consequently, policies
may struggle when objects appear in unfamiliar arrangements
\citep{zhou2025liberopro} or when asked to continue after a failed attempt
\citep{liu2026liberorecover}. Simply retrying the task policy often leaves the
underlying obstruction unresolved. Robots therefore need a complementary
recovery capability that can expand as new failure modes arise.

Learning such recovery behaviors directly on hardware is costly, as exploring
failures requires frequent resets and risks damaging objects
\citep{torne2024rialto,li2026farl}. Reconstructed digital twins provide a
safer and more scalable setting for reproducing failures and testing
corrections \citep{torne2024rialto}, while physical experience remains
essential for verifying and refining them under real contacts and
observations. Effective recovery therefore requires a loop that discovers
behaviors in simulation, validates them on the robot, and feeds physical
experience back into learning.

Language-model agents are natural candidates to coordinate this loop. Recent
models combine multimodal understanding with the ability to use tools,
execute code, and revise solutions from feedback
\citep{anthropic2026opus55,openai2026gpt6astra}. Prior robotics systems show
that agents can interpret failures and develop reusable programs
\citep{liu2023reflect,liang2023cap,lu2026aspire}. The central challenge is to
connect these abilities into a complete recovery process: diagnose a failure,
select or develop a correction, verify that it worked, retain successful
solutions, and use the resulting data to improve the appropriate policy. The
agent must also decide whether new experience should become a reusable
program, a task-policy demonstration, or a recovery-policy demonstration.
When no suitable recovery is available, human demonstrations enter the loop
to resolve the failure and expand future recovery capabilities
\citep{liu2022sirius}.

We present \method (\cref{fig:teaser}), an agent-guided framework that
develops task execution and scene recovery in a reconstructed digital twin,
then continually improves both through real-world experience. In the twin,
the agent explores task execution, diagnoses failures, and develops corrective
behaviors. Successful task and recovery rollouts initialize separate task and
recovery policies, while programs that compose perception and motion
primitives capture reusable recovery strategies. Keeping these data pathways
separate allows the task and recovery policies to specialize in their
distinct roles.

During real-world deployment, the agent monitors task progress, invokes an
applicable learned or programmatic recovery, verifies that the scene is
workable again, and resumes execution. When no suitable recovery succeeds, it
requests human assistance. Successful task rollouts and human task
demonstrations improve the task policy, while human recovery demonstrations
improve the recovery policy. A parallel collection harness runs this loop
across four robot workstations, allowing task and recovery experience to
accumulate concurrently. Newly encountered failure modes also guide the
development of additional skills. Human intervention therefore resolves the
current failure while helping the robot recover more autonomously in the
future.

Our contributions are:
\begin{itemize}
    \item An agent-guided real-to-sim-to-real framework for jointly developing
    task execution and failure recovery. It uses a reconstructed digital twin
    to safely discover corrective behaviors, then verifies and refines them
    through physical robot experience.

    \item A parallel deployment-and-learning loop that coordinates task and
    recovery policies, reusable recovery programs, outcome verification, and
    human demonstrations. Across four concurrently operating robot
    workstations, collected experience improves mean task success by 53.7
    percentage points to 77.5\%. On one task, the observed human-intervention
    rate falls from 87.5\% in the first collection round to 0\% in the fourth.

    \item A comprehensive evaluation on two simulation benchmarks and four
    real-robot tasks. \method exceeds the strongest baseline in our
    comparison by 7.1 percentage points on LIBERO-Pro and 26.9 points on
    MolmoSpaces. Recovery skills provide a further 10-point gain on the
    real-robot tasks, reaching 87.5\% mean success.
\end{itemize}

\section{Related Work}
\label{sec:related}

\vspace{\paraspacebefore}
\paragraph{Robot foundation models and agents.}
Expressive imitation-learning policies \citep{chi2023diffusion,zhao2023aloha}
and large, diverse datasets, including cross-embodiment robot data and
egocentric human video \citep{brohan2022rt1,oxe2023,cai2025innon}, have
produced generalist manipulation policies, including vision-language-action
(VLA) models that map images and language instructions to actions
\citep{brohan2023rt2,octo2024,kim2024openvla,black2024pi0,pertsch2025fast,pi2025pi05,nvidia2025gr00t,fang2026molmoact2}.
Language models ground plans in robot affordances \citep{ahn2022saycan},
replan from closed-loop feedback \citep{huang2022innermonologue}, write
plans, policy code, value maps, and rewards
\citep{singh2022progprompt,liang2023cap,huang2023voxposer,ma2023eureka},
and accumulate reusable skills \citep{wang2023voyager}. A task-management
vision-language model can also direct a VLA over long horizons by replanning
subtasks and visual traces \citep{liu2026lohomanip}. For manipulation, CaP-X
benchmarks coding agents \citep{fu2026capx}, RATs learns skills through
self-directed play \citep{zhang2026rats}, and ASPIRE refines robot programs,
accumulates validated skills, and transfers simulation experience to
real-robot programming \citep{lu2026aspire}.
\method coordinates learned and programmed recoveries around a task policy,
with human demonstrations for refinement.
\vspace{\paraspaceafter}

\vspace{\paraspacebefore}
\paragraph{Failure detection and recovery.}
Prior work explains manipulation failures from summarized experience
\citep{liu2023reflect}, detects and analyzes them with fine-tuned
vision-language models \citep{duan2024aha,ye2025robofac}, monitors execution
against constraints written by a language or vision-language model
\citep{guo2023doremi,zhou2024codeasmonitor}, or predicts them from policy
features, uncertainty, or learned dynamics
\citep{gu2025safe,xu2025fiper,liu2023runtime}. Runtime steering filters a
policy's proposed actions with a vision-language model before they execute
\citep{wu2025forewarn}, and neuro-symbolic methods combine failure detection
with recovery planning \citep{cornelio2024recover}.
Separate recovery controllers have substantial precedent. Recovery RL
separates task reward from constraint satisfaction
\citep{thananjeyan2021recoveryrl}; RecoveryChaining learns to return to states
from which nominal controllers can complete the task and demonstrates
sim-to-real transfer \citep{vats2024recoverychaining}. FARL keeps its offline
recovery policy fixed during online task-policy learning
\citep{li2026farl}. RACER guides task execution and recovery through a shared
visuomotor actor
\citep{dai2024racer}, while FailSafe generates and verifies corrective-action
data for a VLM that assists VLAs \citep{lin2025failsafe}. ReSYNC expands
recovery skills and planning abstractions from failures
\citep{li2026resync}. \method connects recovery development in a reconstructed
scene with visual restoration checks, learned and programmed executors,
and human demonstrations for separate policy updates during deployment.
Benchmarks evaluate perturbation robustness
\citep{li2024simpler,liu2023libero,zhou2025liberopro,kim2026molmospaces} and
recovery from failed executions \citep{liu2026liberorecover}; our protocol
also tracks human intervention alongside task and recovery outcomes.
\vspace{\paraspaceafter}

\vspace{\paraspacebefore}
\paragraph{Learning from deployment and simulation.}
DAgger aggregates expert labels on the states a learner visits
\citep{ross2011dagger}. Human-gated variants let an operator take over
\citep{kelly2018hgdagger,mandlekar2020iwr}, and robot-gated variants request
help based on uncertainty, novelty, risk, or predicted action discrepancy,
including across robot fleets
\citep{menda2018ensembledagger,hoque2021lazydagger,hoque2021thriftydagger,hoque2022fleetdagger}.
Deployed robots also improve from human interventions
\citep{liu2022sirius,liu2024siriusfleet}, language corrections
\citep{shi2024yay,liu2023olaf}, human-in-the-loop reinforcement learning
\citep{luo2024hilserl}, and autonomously collected experience
\citep{bousmalis2023robocat,zhou2024soar,ghasemipour2025selfimproving}.
ENPIRE lets agents manage scene resets, rollouts, verification, and policy
improvement on real robots \citep{xiao2026enpire}.
Before deployment, real-to-sim pipelines build digital twins or cousins of a
scene \citep{torne2024rialto,dai2024acdc} and reconstruct articulated assets
\citep{chen2024urdformer,mandi2024real2code}, while generative pipelines
propose new simulation tasks with demonstrations
\citep{wang2023gensim,wang2023robogen} or synthesize demonstrations from a few
human examples \citep{mandlekar2023mimicgen}. \method combines both:
simulation initializes task and recovery behaviors, and deployment routes task
successes, task demonstrations, and recovery demonstrations to the policy each
one teaches, requesting human input only at failures rather than expert
labels at every visited state.
\vspace{\paraspaceafter}

\section{Method}
\label{sec:method}

\begin{figure}[t]
    \centering
    \includegraphics[width=\linewidth]{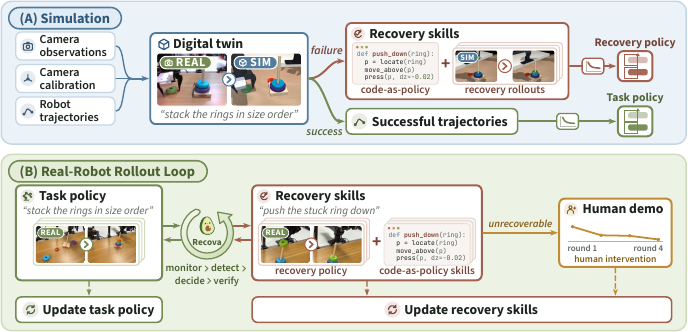}
    \vspace{\capspacebefore}
    \caption{
        \textbf{(A) Simulation:} \method{} builds a digital twin from camera observations, calibration, and robot trajectories. Successful task and recovery rollouts train separate policies, while recovery programs populate a code-as-policy skill library.
        \textbf{(B) Real-robot rollout:} \method{} monitors execution, invokes recovery policies or programs, verifies restoration, and requests human demonstrations when needed. Task successes, verified recoveries, and human demonstrations update the corresponding policies, while new failures guide skill-library expansion. Paired frames show rollout endpoints; the inset tracks human intervention across DAgger rounds.
    }
    \label{fig:pipeline_overview}
    \vspace{\capspaceafter}
\end{figure}

\method decouples task execution from scene recovery and connects the two
through an agent that coordinates simulation, deployment, and learning
(\cref{fig:pipeline_overview}). This separation trains each capability on the
data suited to its role and lets recovery skills accumulate independently of
any single task. \Cref{sec:formulation} formalizes task execution and
recovery. \Cref{sec:simprep} describes how a coding agent builds a digital
twin of the workstation and turns simulated successes and failures into
training data and reusable recovery programs. \Cref{sec:online} presents the
real-robot loop, in which a monitor tracks the task policy, triggers and
verifies recoveries, and involves a human operator only when autonomous
execution falls short. \Cref{sec:learning} describes how each deployment
round refines the task and recovery policies using their respective
experience, while newly encountered failures guide the development of
additional recovery skills.

\subsection{Task Execution and Scene Recovery}
\label{sec:formulation}

Let $g$ be a language instruction for a manipulation task and $o_t$ the camera
observations and robot state at time $t$. A task policy
$\pi_\theta(a_t\mid o_t,g)$ acts until the scene satisfies a completion
condition $C_g$.
The task can often still be completed after a failure, but the task policy may be unable to continue until the scene is corrected. For example, when stacking rings on a peg in size order, the robot may release a ring at an angle, causing it to catch on the peg's tip rather than slide into place.
\method{} treats recovery as a distinct objective that targets the failure
blocking task progress. A recovery may restore a workable scene for the task
policy to resume, or resolve the remaining task through a corrective action.

A recovery skill is specified by a natural-language recovery instruction $z$,
such as ``push the stuck ring down,'' and is executed either by a
language-conditioned recovery policy $\rho_\phi(a_t\mid o_t,z)$ or by a
code-as-policy program from a skill library that composes scene queries and
motion primitives \citep{liang2023cap}. The recovery skill
set $\mathcal{K}$ contains the recovery instructions known for a task.
Conditioning recovery on $z$ rather than on $g$ decouples recovery skills from
individual tasks, so a skill can be reused wherever the same failure recurs.
When neither the task policy nor a recovery skill can make progress, a human
operator provides a demonstration, which becomes training data
(\cref{sec:learning}).

Recovery skills serve three complementary roles, and a single skill may serve
several. \emph{Restorative} skills return the scene to a workable state, such
as uprighting a fallen object. \emph{Enabling} skills prepare the next step,
such as moving an object within reach. \emph{Adaptive} skills change the
approach rather than repeating a failed attempt, such as re-grasping with a
new grasp pose or opening an oven from another direction
(\cref{fig:sim_recovery}).

\subsection{Simulation in a Digital Twin}
\label{sec:simprep}

\vspace{\paraspacebefore}
\paragraph{Agent-built digital twin.}
Exploring failures on hardware requires frequent human resets and risks
damaging objects. \method instead explores each task and its failure modes in
a digital twin built by a coding agent, a language model that writes,
executes, and revises code from execution feedback. The agent reconstructs
the workstation from real-robot recordings: camera observations, robot
trajectories, and camera calibration when available. In our implementation,
the twin is a MuJoCo scene \citep{todorov2012mujoco}. Starting from models of
the robot and the task objects, the agent fits object poses and contact
parameters so that replaying a recorded trajectory in the twin aligns with the
real camera views. This grounds the twin in the real workstation's scale and
layout; \cref{fig:pipeline_overview}(A) pairs a real frame with the twin's
render of the same moment.
\vspace{\paraspaceafter}

\vspace{\paraspacebefore}
\paragraph{Discovering task and recovery behaviors.}
Given $g$, the agent attempts the task in the twin, and successful attempts
form the task dataset. When an attempt fails, the agent diagnoses the failure,
names a recovery instruction $z$, and develops the corresponding skill in two
complementary forms: a program for the skill library, and recovery rollouts
that pair the disturbed scene and $z$ with corrective actions, which form the
recovery dataset.
These datasets provide initial training data for $\pi_\theta$ and $\rho_\phi$,
and the new instructions seed $\mathcal{K}$, so the robot enters deployment
with recovery skills already in place. \method{} queries a VLM agent to assess the current scene and verify whether the recovery policy has restored it to a state from which the task policy can resume (\cref{sec:online}).
\vspace{\paraspaceafter}

\subsection{Real-Robot Rollout Loop}
\label{sec:online}

Simulation gives both policies a strong starting point, but the real
workstation still presents contact dynamics, sensing conditions, and failure
modes that the twin cannot fully anticipate. The real-robot rollout loop
closes this gap: the task policy runs autonomously, \method intervenes only
when needed, and the resulting experience becomes training data that further
improves both policies (\cref{sec:learning}). A vision-language monitor
implements the four steps of
\cref{fig:pipeline_overview}(B), and \cref{alg:orchestration} in
\cref{app:orchestration} gives the complete control flow.

\begin{itemize}
    \item \textbf{Monitor.} Before the first attempt, a requirement query converts
    $g$ and the initial images into an explicit completion condition $C_g$.
    While the task policy acts, a completion query periodically checks whether
    $C_g$ holds.
    \item \textbf{Detect.} At a finer interval, an intervention query
    compares recent frames with the initial scene to decide
    whether to intervene. It also reports whether the scene is still intact
    and names a recovery instruction $z$, reusing an existing instruction from
    $\mathcal{K}$ whenever one applies so that demonstrations of the same
    behavior share a label.
    \item \textbf{Decide.} If the scene is intact, the task policy has stalled
    without disturbing it, and the operator provides a task demonstration.
    Otherwise, the recovery policy executes $z$ within a fixed time budget if $z$ is in its
    registry of trained instructions. If no learned skill applies or the attempt fails, the
    operator demonstrates the recovery, which is recorded under $z$. After two
    consecutive interventions on a task, the next attempt is a full human
    demonstration, which avoids cycles of repeated failure.
    \item \textbf{Verify.} After an autonomous recovery, a restoration query uses the initial images as a reference to check whether the scene is ready for another task attempt. Objects do not need to return to their exact original positions and orientations. Once recovery is verified, the robot returns to its home pose and the task policy starts a new attempt.
\end{itemize}

Because human assistance is needed only at these handoffs, one operator can supervise multiple stations with different scene configurations in parallel (\cref{fig:collection}). This scales data collection without requiring additional operators. On real robots, the learned recovery policy performs autonomous recoveries. Meanwhile, the agent develops and tests new recovery programs in the digital twin, expanding the skill library (\cref{sec:simprep}).

\subsection{Learning Across DAgger Rounds}
\label{sec:learning}

\begin{figure}[t]
    \centering
    \includegraphics[width=\linewidth]{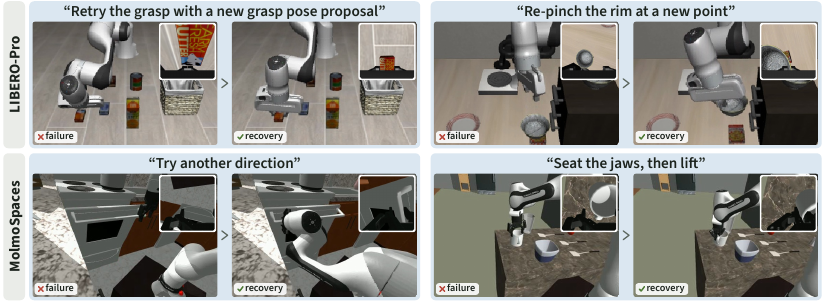}
    \vspace{\capspacebefore}
    \caption{\textbf{Recovery skills in simulation.} The failure (left) and
    recovery (right) moments of one LIBERO-Pro (top) or MolmoSpaces (bottom)
    episode: the main camera view, with the wrist camera inset at the top
    right, under the recovery instruction.}
    \label{fig:sim_recovery}
    \vspace{\capspaceafter}
\end{figure}
\begin{table}[t]
  \centering
  \vspace{\capspacebefore}
  \caption{\textbf{MolmoSpaces task success (\%).} P\&P denotes pick-and-place.
  Avg. is the unweighted mean across the four simulation task categories.
  Bold marks the best reported value in each column. Baseline rows reproduce
  published results.}
  \label{tab:molmospaces}
  \vspace{\capspaceafter}
  \small
  \setlength{\aboverulesep}{0pt}\setlength{\belowrulesep}{0pt}
  \setlength{\extrarowheight}{2.4pt}\setlength{\tabcolsep}{5pt}
  \begin{tabularx}{\linewidth}{l*{5}{Y}}
    \toprule
    & \multicolumn{5}{c}{\textbf{MolmoSpaces}\,$\uparrow$} \\
    \cmidrule(l){2-6}
    Methods & Pick & P\&P & Open & Close & \textbf{Avg.} \\
    \midrule
    \rowcolor{tablefamily}\multicolumn{6}{l}{\textit{Agentic code-as-policy methods}} \\
    CaP-Agent0~\citep{fu2026capx} & 23.0 & 11.0 & 14.0 & 36.0 & 21.0 \\
    RATs~\citep{zhang2026rats} & 37.0 & 22.0 & 20.0 & 73.0 & 38.0 \\
    \midrule
    \rowcolor{tablefamily}\multicolumn{6}{l}{\textit{Vision-language-action policies}} \\
    $\pi_0$~\citep{black2024pi0} & 14.0 & 7.0 & 10.0 & 47.0 & 19.5 \\
    $\pi_0$-FAST~\citep{pertsch2025fast} & 20.0 & 10.0 & 9.0 & 35.0 & 18.5 \\
    $\pi_{0.5}$~\citep{pi2025pi05} & 28.0 & 10.0 & 18.0 & 62.0 & 29.5 \\
    MolmoAct2-DROID~\citep{fang2026molmoact2} & 41.0 & 32.0 & 3.8 & 67.0 & 36.0 \\
    \midrule
    \rowcolor{tableours}\textbf{\method (ours)} & \textbf{78.7} & \textbf{40.5} & \textbf{56.5}
      & \textbf{83.9} & \textbf{64.9} \\
    \bottomrule
  \end{tabularx}
  \vspace{\tabspacebelow}
\end{table}

\method{} refines the task and recovery policies through repeated rounds of DAgger-style data collection \citep{ross2011dagger}. During round $k$, both policies remain fixed while the agent collects trajectories from robot execution and human demonstrations. These trajectories are divided into four groups. $\mathcal{S}_k$ (\texttt{Success}) contains task rollouts completed without human takeover, and $\mathcal{E}_k$ (\texttt{Expand}) contains human task demonstrations. $\mathcal{R}_k$ (\texttt{Recover}) contains human recovery demonstrations, each labeled with its recovery instruction $z$, while $\mathcal{A}_k$ contains autonomous recoveries that pass the scene-restoration check. Failed policy segments are saved for analysis but excluded from training.

At the end of each round, \method{} adds successful task rollouts and human task demonstrations to the task dataset, and human recovery demonstrations and verified autonomous recoveries to the recovery dataset:
\begin{align}
\mathcal{D}^{k+1}_{\mathrm{task}}
  &= \mathcal{D}^{k}_{\mathrm{task}} \cup \mathcal{S}_k \cup \mathcal{E}_k,
  &\mathcal{D}^{k+1}_{\mathrm{rec}}
  &= \mathcal{D}^{k}_{\mathrm{rec}} \cup \mathcal{R}_k \cup \mathcal{A}_k.
\label{eq:aggregation}
\end{align}
Each policy is then fine-tuned on its updated dataset using its standard
supervised objective, with the task policy conditioned on the task
instruction $g$ and the recovery policy on the recovery instruction $z$.
Keeping these datasets separate allows each policy to learn the behavior it is responsible for: completing the task or restoring a scene from which task execution can resume.

The recovery skill set also expands as the robot encounters new failures. When no known recovery instruction applies, the intervention query proposes a new instruction, and a human demonstrates the corresponding recovery. This demonstration provides the first training example for the new skill. The instruction is added to $\mathcal{K}$ and guides the development of recovery programs in the digital twin (\cref{sec:simprep}). After the recovery policy is trained on the new demonstrations, the instruction is added to the recovery policy's registry, making the
learned skill available for autonomous execution. In this way, human interventions provide training examples that can help the robot handle similar failures autonomously in later rounds.

\section{Experiments}
\label{sec:experiments}

\begin{table}[t]
  \centering
  \vspace{\capspacebefore}
  \caption{\textbf{LIBERO-Pro task success (\%).} Object, Goal, and Spatial
  each include initial-position swaps (Pos.) and task perturbations (Task).
  Avg. is the unweighted mean of all six columns. Bold marks the best
  reported value in each column. Baseline rows reproduce published results.}
  \label{tab:libero}
  \vspace{\capspaceafter}
  \small
  \setlength{\aboverulesep}{0pt}\setlength{\belowrulesep}{0pt}
  \setlength{\extrarowheight}{2.4pt}\setlength{\tabcolsep}{5pt}
  \begin{tabularx}{\linewidth}{l*{7}{Y}}
    \toprule
    & \multicolumn{2}{c}{\textbf{Object}\,$\uparrow$}
    & \multicolumn{2}{c}{\textbf{Goal}\,$\uparrow$}
    & \multicolumn{2}{c}{\textbf{Spatial}\,$\uparrow$} & \\
    \cmidrule(lr){2-3}\cmidrule(lr){4-5}\cmidrule(lr){6-7}
    Methods & Pos. & Task & Pos. & Task & Pos. & Task & \textbf{Avg.}\,$\uparrow$ \\
    \midrule
    \rowcolor{tablefamily}\multicolumn{8}{l}{\textit{Agentic code-as-policy methods}} \\
    CaP-Agent0~\citep{fu2026capx} & 27.0 & 31.0 & 29.0 & 16.0 & 13.0 & 23.0 & 23.2 \\
    RATs~\citep{zhang2026rats} & 61.0 & 63.0 & 43.0 & 36.0 & 29.0 & 31.0 & 43.8 \\
    ASPIRE~\citep{lu2026aspire} & \textbf{98.0} & \textbf{95.0} & 81.0 & 45.0 & 51.0 & 60.0 & 71.7 \\
    \midrule
    \rowcolor{tablefamily}\multicolumn{8}{l}{\textit{Vision-language-action policies}} \\
    $\pi_0$~\citep{black2024pi0} & 0.0 & 0.0 & 0.0 & 0.0 & 0.0 & 0.0 & 0.0 \\
    $\pi_{0.5}$~\citep{pi2025pi05} & 17.0 & 1.0 & 38.0 & 0.0 & 20.0 & 1.0 & 12.8 \\
    MolmoAct2-LIBERO~\citep{fang2026molmoact2} & 52.0 & 24.0 & 20.0 & 21.0 & 27.0 & 57.0 & 33.5 \\
    \midrule
    \rowcolor{tableours}\textbf{\method (ours)} & 91.2 & 87.2 & \textbf{82.2} & \textbf{67.0}
      & \textbf{62.2} & \textbf{83.0} & \textbf{78.8} \\
    \bottomrule
  \end{tabularx}
  \vspace{\tabspacebelow}
\end{table}

We evaluate \method in simulation and on real robots around three questions.
(1) Does \method raise task success under perturbations over code-as-policy
agents and VLA policies? (2) Which recovery skills does it develop and
execute? (3) Does deployment with \method reduce human assistance over
successive rounds?
\Cref{sec:simexp} addresses the first question on LIBERO-Pro and
MolmoSpaces. \Cref{sec:realresults} evaluates task and recovery behavior on
four real-robot tasks and examines human intervention during data collection.

\subsection{Simulation Benchmarks and Comparisons}
\label{sec:simexp}

\vspace{\paraspacebefore}
\paragraph{Benchmarks and baselines.}
We evaluate on two simulation benchmarks that test manipulation beyond nominal configurations.
Our LIBERO-Pro evaluation \citep{zhou2025liberopro} covers the Object, Goal, and Spatial suites under initial-position swaps and task perturbations, yielding six settings. Position swaps exchange object locations, while task perturbations change the instruction to target another object. These settings test adaptation to changes in both scene layout and task requirements. Following prior work \citep{lu2026aspire}, we evaluate each task on 50 initial states.
MolmoSpaces \citep{kim2026molmospaces} spans four manipulation categories across diverse simulated scenes: pick, pick-and-place, open, and close. These categories cover rigid-object grasping and placement as well as articulated-object manipulation, such as opening and closing cabinets or drawers.

We compare against \emph{code-as-policy (CaP) methods}, which use language models to generate executable robot-control programs, and \emph{vision-language-action (VLA) policies}, which map visual observations and language instructions to robot actions.
CaP baselines include CaP-Agent0 \citep{fu2026capx}, RATs \citep{zhang2026rats}, and, on LIBERO-Pro, ASPIRE \citep{lu2026aspire}. CaP-Agent0 uses multi-turn execution feedback to refine control programs, RATs acquires reusable code skills through self-directed play, and ASPIRE expands a skill library through iterative program refinement and exploration. VLA baselines include $\pi_0$ \citep{black2024pi0}, $\pi_{0.5}$ \citep{pi2025pi05}, the MolmoAct2-LIBERO and MolmoAct2-DROID checkpoints \citep{fang2026molmoact2}, and, on MolmoSpaces, $\pi_0$-FAST \citep{pertsch2025fast}. Baseline results are taken from prior publications, while \method{} results are obtained from our own evaluation.
\vspace{\paraspaceafter}

\vspace{\paraspacebefore}
\paragraph{Benchmark implementation.}
In the simulation benchmarks, \method{} pairs a MolmoAct2 task policy
\citep{fang2026molmoact2} with programmatic recovery skills. A recovery
program takes over from the state left by the policy when its execution
budget is exhausted or, in LIBERO-Pro, progress stalls. Program development
and evaluation use disjoint random seeds, with held-out initial object
configurations at evaluation. Task success is the fraction of evaluation
trials satisfying the task's completion criterion, including cases where
the recovery program completes the task.
\vspace{\paraspaceafter}

\vspace{\paraspacebefore}
\paragraph{LIBERO-Pro results.}
\method{} achieves the highest reported mean success rate of 78.8\% across six settings, exceeding ASPIRE, the strongest baseline on average, by 7.1 percentage points (\Cref{tab:libero}). The largest gains occur under task perturbations in the Goal and Spatial suites, where \method{} outperforms ASPIRE by 22.0 and 23.0 percentage points, respectively. Although ASPIRE remains stronger on the Object suite, \method{} leads in all four Goal and Spatial settings, suggesting particular strength in handling the task variations evaluated in these suites.
\vspace{\paraspaceafter}

\vspace{\paraspacebefore}
\paragraph{MolmoSpaces results.}
\method{} achieves the highest reported success rate in all four MolmoSpaces categories, with a mean of 64.9\%, exceeding RATs, the strongest baseline on average, by 26.9 percentage points (\Cref{tab:molmospaces}). Consistent gains across picking, pick-and-place, opening, and closing suggest that the framework's benefits extend across diverse manipulation behaviors. Pick-and-place remains the most challenging category, but \method{} reaches 40.5\% success, compared with 32.0\% for MolmoAct2-DROID, the strongest baseline in this category. Qualitative examples in \Cref{fig:sim_recovery} illustrate how recovery behaviors address execution failures across both benchmarks.
\vspace{\paraspaceafter}

\subsection{Real-Robot Experiments}
\label{sec:realresults}
\label{sec:real}

\begin{figure}[t]
    \centering
    \includegraphics[width=\linewidth]{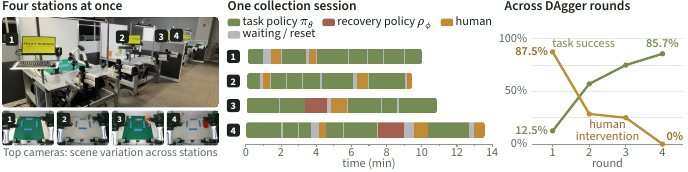}
    \vspace{\capspacebefore}
    \caption{\textbf{Parallel automated DAgger data collection.}
    \textbf{Left:} Four robot stations operate in parallel with varied scene configurations.
    \textbf{Middle:} Station activity during one collection session, distinguishing task-policy execution, recovery-policy execution, human control, and waiting or reset periods.
    \textbf{Right:} Observed task success and human intervention rates across four collection rounds. As policies are fine-tuned on accumulated data, task success increases while human intervention decreases.}
    \label{fig:collection}
    \vspace{\capspaceafter}
\end{figure}

\vspace{\paraspacebefore}
\paragraph{Setup.}
We evaluate four tasks on bimanual YAM workstations: packing a pencil box,
stacking rings, and drawing and discarding a mahjong tile. Each workstation
has two I2RT YAM arms with parallel grippers, four Intel RealSense cameras
(overhead, front, and one on each wrist), and a 30\,Hz control loop. For
each task, the task and recovery policies are fine-tuned from the
$\pi_{0.5}$ base checkpoint \citep{pi2025pi05}; the task policy observes the
overhead and wrist cameras, and the recovery policy is conditioned on the
recovery instruction. Gemini 3.8 Flash serves as the monitor and sees all
four views. In each task's digital twin, the agent developed six or seven
recovery skills, 25 in total (\cref{app:skill_gallery}). Each task is
evaluated over 20 trials per configuration: the base task policy, after DAgger
fine-tuning, and with recovery skills enabled at deployment.
\vspace{\paraspaceafter}

\begin{table}[t]
  \centering
  \vspace{\capspacebefore}
  \caption{\textbf{Real-robot task success (\%)} over 20 trials per task and
  configuration. Rows add DAgger fine-tuning of the task policy, then
  recovery skills at deployment. Bold marks the best result for each task.}
  \label{tab:real}
  \vspace{\capspaceafter}
  \small
  \setlength{\aboverulesep}{0pt}\setlength{\belowrulesep}{0pt}
  \setlength{\extrarowheight}{2.4pt}\setlength{\tabcolsep}{5pt}
  \begin{tabularx}{\linewidth}{l*{5}{Y}}
    \toprule
    Policy & Pencil box & Stack rings & Draw tile & Discard tile & \textbf{Avg.}\,$\uparrow$ \\
    \midrule
    Base task policy & 20.0 & 35.0 & 25.0 & 15.0 & 23.8 \\
    + DAgger & 70.0 & 75.0 & 85.0 & 80.0 & 77.5 \\
    \rowcolor{tableours}\textbf{+ Recovery skills} & \textbf{85.0} & \textbf{90.0} & \textbf{90.0}
      & \textbf{85.0} & \textbf{87.5} \\
    \bottomrule
  \end{tabularx}
  \vspace{\tabspacebelow}
\end{table}

\vspace{\paraspacebefore}
\paragraph{Task success.}
DAgger fine-tuning more than triples the base task policy's mean success,
from 23.8\% to 77.5\% (\cref{tab:real}), and enabling recovery skills at
deployment raises it further to 87.5\%, the best result on every task.
Recovery contributes the largest gains on ring stacking and pencil-box
packing, 15.0 percentage points each, consistent with tasks in which a single
misplaced ring or pencil blocks the remaining steps. On the two mahjong tasks,
whose DAgger policies already succeed in at least 80\% of trials, recovery
still adds 5.0 points each.
\vspace{\paraspaceafter}

\vspace{\paraspacebefore}
\paragraph{DAgger rounds.}
Beyond task success, we track the human intervention rate, the fraction of
collection episodes involving human takeover. Collection episodes include
policy attempts and human demonstrations, excluding operator-discarded
episodes. A verified recovery ends its episode, and resumed task execution
begins a new attempt. The right panel of \cref{fig:collection} follows four
DAgger rounds on drawing a mahjong tile (seven or eight episodes each), with
outcomes measured before the collected data are used for the next policy
update. The fraction of collection episodes completed by the task policy
rose from 12.5\% to 85.7\%. Human takeovers occurred in seven of eight
episodes in the first round (87.5\%) and none of seven in the fourth (0\%). The
middle panel shows a later 14-minute session on four stations at once: the
task policy completed 22 of 31 episodes, and human control accounted for
11\% of the summed station time.
\vspace{\paraspaceafter}

\begin{figure}[t]
    \centering
    \includegraphics[width=\linewidth]{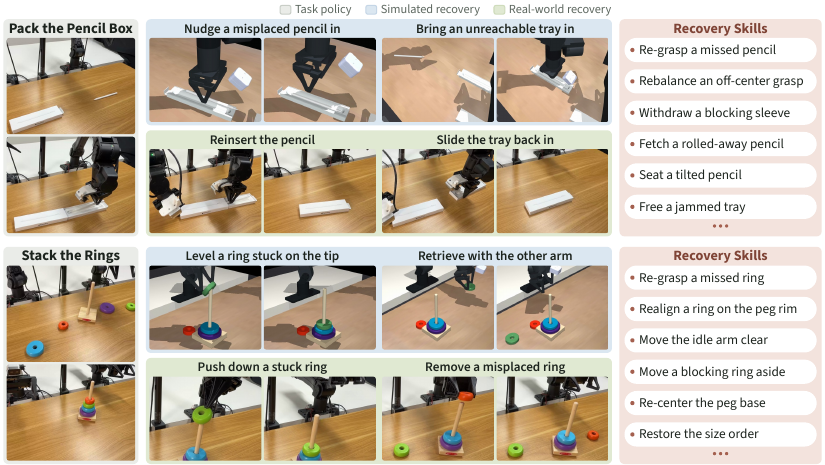}
    \vspace{\capspacebefore}
    \caption{\textbf{Real-robot tasks and their recovery skills.} For each
    task, the gray panel shows the start and result of a task-policy
    rollout. Each recovery skill appears under its name as two frames of one
    rollout, the action and then the result, developed in the digital twin
    (blue) or executed on the robot by a learned recovery policy (green). The
    list on the right names further recovery skills.}
    \label{fig:recovery_skills}
    \vspace{\capspaceafter}
\end{figure}

\vspace{\paraspacebefore}
\paragraph{Recovery skills.}
\Cref{fig:recovery_skills} shows two of the four tasks, each with two skills
developed in the digital twin and two learned recovery policies executed on
the robot. In \emph{ring stacking}, the twin skills address failures such as
a missed pickup or a ring caught on the peg tip, and on the robot, learned
recovery policies push down a stuck ring and remove a ring stacked out of
order. In \emph{pencil-box packing}, they handle missed grasps, a pencil
jammed at the tray, and objects out of reach, and the recovery policies
reinsert the pencil and use both arms to slide a detached tray back into its
sleeve. For the two mahjong tasks, the recovery policies stand a fallen tile
back upright, return a drawn tile to the wall, and realign a displaced hand
tile. \Cref{fig:teaser} shows one real-robot recovery per task.
\vspace{\paraspaceafter}

\section{Conclusion}
\label{sec:conclusion}

We presented \method, an agent-guided framework that treats task execution and
scene recovery as distinct, coordinated capabilities. A coding agent develops
both in a reconstructed digital twin, and a deployment loop monitors the task
policy, invokes and verifies recovery skills, and turns unresolved failures
into demonstrations for the policy they can teach. \method outperforms
code-as-policy agents and VLA policies on two simulation benchmarks, its
recovery skills improve real-robot success beyond DAgger fine-tuning alone,
and the collection study shows declining human intervention over four rounds
on one task.
These results establish scene recovery as a learnable capability that
complements task execution, and show that agents can turn failures into
reusable skills. Automating the registration of newly learned skills and
sharing them across tasks will further expand the set of recovery skills.

\section*{Acknowledgments}
We thank Cristaldo Campos, Jimmy Wu, and Tingwu Wang for their help with the
real robot infrastructure and filming the demo.

\clearpage
\bibliographystyle{plainnat}
\bibliography{references}

\clearpage
\appendix
\raggedbottom
\addtocontents{toc}{\protect\setcounter{tocdepth}{2}}
\renewcommand{\contentsname}{Appendix Table of Contents}
\tableofcontents
\section{Real-Robot Orchestration}
\label{app:orchestration}

\Cref{alg:orchestration} gives the control flow that the real-robot harness
runs in deployment round~$k$ (\cref{sec:online}). The monitor $M$ is
realized as four vision-language-model queries: completion
($M_{\mathrm{done}}$), intervention ($M_{\mathrm{int}}$, which also reports
whether the scene is intact and gives a recovery instruction $z$),
restoration ($M_{\mathrm{res}}$), and requirement generation
($M_{\mathrm{req}}$). $I_0$ and $I_t$ are the initial and current images
from the monitor's four cameras, $H_t$ holds the two most recent snapshots,
taken one second apart, and $V_t$ holds the round's last three intervention
verdicts. $\mathcal{K}$ is the set of known recovery instructions, which the
intervention query reuses verbatim when one applies;
$\mathcal{K}_\rho \subseteq \mathcal{K}$ holds those in the learned-skill
registry, for which the recovery policy $\rho_\phi$ is deployed. Step counts
are control steps at 30\,Hz, with the harness defaults shown.

\begin{algorithm}[!t]
\caption{\method's orchestration in real-robot deployment round $k$.}
\label{alg:orchestration}
\algrenewcommand\algorithmicrequire{\textbf{Input:}}
\algrenewcommand\algorithmicensure{\textbf{Output:}}
\begin{algorithmic}[1]
\Require task $g$ with cached completion requirement $C_g$; policies
  $\pi_\theta$ and $\rho_\phi$; instructions $\mathcal{K}$ and registry
  $\mathcal{K}_\rho \subseteq \mathcal{K}$; check intervals $N_c = 600$ and
  $N_i = 300$; step limits $T_{\max} = 5400$ and $T_{\mathrm{rec}} = 1500$;
  demonstration threshold $c_{\max} = 2$
\Ensure trajectory collections $\mathcal{S}_k$, $\mathcal{E}_k$,
  $\mathcal{R}_k$, and $\mathcal{A}_k$
\State $c \gets 0$ \Comment{consecutive interventions on $g$}
\While{continuous operation is enabled} \Comment{one task attempt per iteration}
  \State return the robot to its home pose
  \If{$c \ge c_{\max}$}
    \State a human demonstrates $g$; add the demonstration to $\mathcal{E}_k$; $c \gets 0$
    \State \textbf{continue}
  \EndIf
  \State $I_0 \gets$ images of the current scene
  \If{$C_g$ is empty}
    \State $C_g \gets M_{\mathrm{req}}(g, I_0)$ with the robot stationary; cache $C_g$
  \EndIf
  \State start recording trajectory $\tau$; $t \gets 0$; $d \gets 0$
  \Repeat
    \State execute $a_t \sim \pi_\theta(\cdot \mid o_t, g)$; $t \gets t + 1$
    \If{$t \bmod N_c = 0$}
      \State $d \gets M_{\mathrm{done}}(g, C_g, I_0, I_t)$ with the robot held
    \EndIf
    \If{$t \bmod N_i = 0$}
      \State submit $M_{\mathrm{int}}(g, C_g, I_0, H_t, V_t, \mathcal{K})$; $\pi_\theta$ keeps running
    \EndIf
    \If{$d = 1$} \Comment{completion takes precedence}
      \State add $\tau$ to $\mathcal{S}_k$; $c \gets 0$
    \ElsIf{an intervention verdict $(\mathit{intact}, z)$ has arrived}
      \State $c \gets c + 1$; discard $\tau$; return the robot to its home pose
      \If{$\mathit{intact}$} \Comment{stalled progress, no disturbance}
        \State a human finishes $g$; add the demonstration to $\mathcal{E}_k$; $c \gets 0$
      \Else
        \State \Call{Recover}{$z$} \Comment{the next attempt resumes $\pi_\theta$}
      \EndIf
    \EndIf
  \Until{the attempt has ended or $t = T_{\max}$} \Comment{$\tau$ is discarded at $T_{\max}$}
\EndWhile
\Statex
\Procedure{Recover}{$z$}
  \If{$z \in \mathcal{K}_\rho$} \Comment{one bounded autonomous attempt}
    \State start recording trajectory $\tau'$
    \For{$s = 1, \dots, T_{\mathrm{rec}}$}
      \State execute $a_s \sim \rho_\phi(\cdot \mid o_s, z)$
      \If{$s \bmod N_i = 0$ or $s = T_{\mathrm{rec}}$}
        \If{$M_{\mathrm{res}}(z, I_0, I_s) = 1$ (robot held)}
          \State add $\tau'$ to $\mathcal{A}_k$ \Comment{verified restoration}
          \State \Return
        \EndIf
      \EndIf
    \EndFor
    \State return the robot to its home pose \Comment{not restored in time}
  \EndIf
  \State a human performs only the recovery; add the segment, labeled $z$, to $\mathcal{R}_k$
\EndProcedure
\end{algorithmic}
\end{algorithm}

The harness also handles details that \cref{alg:orchestration} omits. Every
verdict carries its round and submission step, and verdicts from an ended
round or from before a manual scene adjustment are discarded. When an
intervention verdict arrives while a completion query is still running, the
harness holds the robot until the completion verdict lands. Intervention
queries run asynchronously, so the task policy keeps acting while they are
evaluated. If no recovery
instruction is available, the human segment is labeled ``Restore the scene
to its pre-task state.'' The operator can also pause a rollout to adjust
the scene by hand, which adds no demonstration; take over at any time, which
records a task demonstration in $\mathcal{E}_k$; request a task or recovery
demonstration directly; flip the pending label before it is saved; and stop
or discard the round. A discarded prefix is excluded from the training data
but remains in the session log with the monitor's prompts and verdicts.

\section{Monitor Prompts and Example Verdicts}
\label{app:monitor}

The monitor sends every query with the same system prompt and one of four
templates (\cref{fig:monitor_prompts}). Each image is a $448\times336$ composite
of the four cameras at one moment: top and front views in the upper row and the
two wrist views in the lower row (the prompts call the front camera the bottom
camera). The requirement query sees the four initial views; the completion and
restoration queries see the initial and current composites; and the
intervention query sees the initial composite and the two most recent
snapshots, together with its three previous verdicts and the known recovery
skills for the task. All queries run at temperature 0 and return JSON. Across
the four-station collection sessions, the median latency is 4.5\,s for
intervention, 5.0\,s for completion, and 5.5\,s for restoration queries.
\Cref{fig:monitor_examples} shows recorded verdicts from these sessions.

\definecolor{cardframe}{HTML}{D5DAE2}
\definecolor{cardback}{HTML}{F7F8FA}
\definecolor{cardtitle}{HTML}{2F3747}
\definecolor{cardreply}{HTML}{E9EDF2}
\definecolor{cardlabel}{HTML}{6B7484}
\definecolor{chiptext}{HTML}{DFEAF7}
\definecolor{chipimage}{HTML}{E3F0CF}
\definecolor{placeholder}{HTML}{1F6FB2}
\definecolor{jsonkey}{HTML}{1D6F8A}
\definecolor{jsonstr}{HTML}{A31515}
\definecolor{jsonlit}{HTML}{0B3FB5}
\newcommand{\figsans}{\fontfamily{NVIDIASans}\selectfont}
\newcommand{\figmono}{\fontfamily{JetBrainsMono}\fontsize{6.2}{7.7}\selectfont}
\newcommand{\codeblock}{\raggedright\hyphenpenalty=10000\exhyphenpenalty=10000\parindent=0pt\relax}
\newcommand{\ph}[1]{\textcolor{placeholder}{#1}}
\newcommand{\jk}[1]{\textcolor{jsonkey}{#1}}
\newcommand{\js}[1]{\textcolor{jsonstr}{#1}}
\newcommand{\jl}[1]{\textcolor{jsonlit}{#1}}
\newcommand{\cardlabel}[1]{{\figsans\fontseries{b}\fontsize{6.5}{8}\selectfont\textcolor{cardlabel}{#1}}}
\newcommand{\chip}[2]{\tcbox[on line, colback=#1, colframe=#1, boxrule=0pt, arc=1.5pt,
  boxsep=0pt, left=2.5pt, right=2.5pt, top=1pt, bottom=1pt,
  fontupper=\figsans\fontsize{6.5}{7.5}\selectfont]{#2}}
\newcommand{\textin}[1]{\chip{chiptext}{#1}}
\newcommand{\imagein}[1]{\chip{chipimage}{#1}}
\newcommand{\chipsep}{\hskip 2pt plus 1pt\relax}
\newcommand{\inputs}[2]{\cardlabel{Text}\enspace#1\hskip 5pt plus 2pt\cardlabel{Images}\enspace#2\par\vspace{2pt}}
\newcommand{\inputline}[2]{\par\hangindent=24pt\hangafter=1\noindent\makebox[24pt][l]{\cardlabel{#1}}#2\par}
\newcommand{\stackedinputs}[2]{\inputline{Text}{#1}\inputline{Images}{#2}\vspace{2pt}}
\newcommand{\reply}{\cardlabel{Reply}\enspace}
\newlength{\monimgw}
\newlength{\monimgh}
\setlength{\monimgw}{0.232\linewidth}
\setlength{\monimgh}{0.75\monimgw}
\tcbset{
  monitorbase/.style={arc=1.5pt, boxrule=0.5pt, colframe=cardframe, colback=cardback,
    colbacktitle=cardtitle, coltitle=white,
    fonttitle={\figsans\fontseries{b}\fontsize{8}{9.5}\selectfont},
    toptitle=1.2pt, bottomtitle=1.2pt, lefttitle=5pt, righttitle=5pt,
    left=5pt, right=5pt, top=3pt, bottom=3pt, fontupper={\figmono\codeblock}},
  promptcard/.style={monitorbase, bicolor, colbacklower=cardreply, segmentation hidden,
    middle=2pt, fontlower={\figmono\codeblock}, before skip=0pt, after skip=0pt},
  systemcard/.style={monitorbase, bicolor, sidebyside, sidebyside align=center,
    lefthand width=57pt, sidebyside gap=10pt, boxsep=0.5mm,
    colback=cardtitle, colbacklower=cardback, segmentation hidden,
    fontupper={\figsans\fontseries{b}\fontsize{7.6}{9}\selectfont\color{white}\raggedright},
    fontlower={\figmono\codeblock}, before skip=0pt, after skip=3pt},
  verdictcard/.style={monitorbase, enhanced, height=\monimgh, valign=center, nobeforeafter,
    fontupper={\fontfamily{JetBrainsMono}\fontsize{6.4}{7.9}\selectfont\codeblock}}}

\newcommand{\requirementcard}[1]{%
\begin{tcolorbox}[promptcard, title={Requirement query\hfill$M_{\mathrm{req}}$}, #1]
\inputs{\textin{Task instruction: {\figmono"\ph{\{instruction\}}"}}}{\imagein{four initial views}}
Write the completion requirement for this task: a concise, objectively checkable
description of what the scene must look like once the task has been fully
completed (object positions/arrangement, gripper state, anything that must NOT
have changed). 1-3 sentences, only about the final scene state, verifiable
purely from camera images.
\tcblower
\reply\{\jk{"requirement"}: \js{"<the requirement>"}\}
\end{tcolorbox}}
\newcommand{\completioncard}[1]{%
\begin{tcolorbox}[promptcard, title={Completion query\hfill$M_{\mathrm{done}}$}, #1]
\stackedinputs{\textin{Task instruction}\chipsep\textin{completion requirement}}%
  {\imagein{initial and current composites}}
Has the task been fully completed? Compare against the initial scene to judge
what has changed. Be strict: answer true only when the requirement is clearly
and completely satisfied; partial progress or an ambiguous view is false. Check
EVERY moved object: if any object left its initial position but sits at no
valid goal position, the task is NOT complete.
\tcblower
\reply\{\jk{"complete"}: \jl{true}/\jl{false}, \jk{"reason"}: \js{"<one short sentence>"}\}
\end{tcolorbox}}
\newcommand{\interventioncard}[1]{%
\begin{tcolorbox}[promptcard, title={Intervention query\hfill$M_{\mathrm{int}}$}, #1]
\stackedinputs{\textin{Task instruction}\chipsep\textin{completion requirement}\chipsep
  \textin{operator notes}\chipsep\textin{previous verdicts}}%
  {\imagein{initial composite}\chipsep\imagein{snapshots 1\,s ago and now}}
Decide whether a HUMAN needs to take over NOW. Set intervene = true only when
the snapshots clearly show one of:
\begin{itemize}[nosep, leftmargin=1.6em, labelsep=0.6em, label=-]
\item an object accidentally dropped, fallen, or knocked over;
\item the robot stuck on ONE atomic action for more than \textasciitilde30 s (use your previous verdicts);
\item an EXTRA object moved by mistake lies loose (neither at its start nor at a goal);
\item for hand-tile tasks: a hand tile standing tilted or not facing forward;
\item the scene disturbed so badly the task cannot proceed.
\end{itemize}
A deliberate intermediate placement is NOT a failure. When intervene is true,
set scene\_intact = true if nothing is displaced (the robot merely failed to
make progress; recovery is empty), or false with ONE short imperative recovery
instruction. Known recovery skills: \ph{<list>}. When a listed skill matches,
return it VERBATIM; otherwise write a new one.
\tcblower
\reply\{\jk{"intervene"}: \jl{true}/\jl{false}, \jk{"reason"}: \js{"..."},
\jk{"scene\_intact"}: \jl{true}/\jl{false}, \jk{"recovery"}: \js{"<one imperative clause or empty>"}\}
\end{tcolorbox}}
\newlength{\promptcolw}
\setlength{\promptcolw}{\dimexpr0.5\linewidth-1.5pt\relax}
\newcommand{\promptcol}[1]{\begin{minipage}[t]{\promptcolw}\vspace{0pt}#1\end{minipage}}
\newsavebox{\promptleft}
\newsavebox{\promptcomp}
\newsavebox{\promptright}
\newlength{\promptcolh}
\newlength{\promptcomph}

\begin{figure}[!t]
\centering
\begin{tcolorbox}[systemcard]
System prompt\par{\mdseries\fontsize{6.2}{7.5}\selectfont for every query}
\tcblower
You are a robot-task monitor. You watch a dual-arm robot manipulation rollout
through four cameras (top, bottom, and one on each wrist) and judge its status.
Respond ONLY with a JSON object, no markdown fences.
\end{tcolorbox}
\sbox{\promptleft}{\promptcol{\requirementcard{after skip=3pt}\completioncard{}}}%
\sbox{\promptcomp}{\promptcol{\completioncard{}}}%
\sbox{\promptright}{\promptcol{\interventioncard{}}}%
\setlength{\promptcolh}{\dimexpr\ht\promptleft+\dp\promptleft\relax}%
\ifdim\dimexpr\ht\promptright+\dp\promptright\relax>\promptcolh
  \setlength{\promptcolh}{\dimexpr\ht\promptright+\dp\promptright\relax}%
\fi
\setlength{\promptcomph}{\dimexpr\promptcolh-\ht\promptleft-\dp\promptleft+\ht\promptcomp+\dp\promptcomp\relax}%
\promptcol{\requirementcard{after skip=3pt}%
  \completioncard{height from={\promptcomph} to {\textheight}, space to upper}}\hfill
\promptcol{\interventioncard{height from={\promptcolh} to {\textheight}, space to upper}}
\par\vspace{3pt}
\begin{tcolorbox}[promptcard, title={Restoration query\hfill$M_{\mathrm{res}}$}]
\inputs{\textin{Recovery instruction: {\figmono"\ph{\{recovery\}}"}}}{\imagein{initial and current composites}}
Has the recovery been completed? The INITIAL images define what a valid
pre-task state looks like. If the current arrangement matches it or is an
equivalently workable starting state, the scene IS restored. Exact object
positions need not match.
\tcblower
\reply\{\jk{"restored"}: \jl{true}/\jl{false}, \jk{"reason"}: \js{"<one short sentence>"}\}
\end{tcolorbox}
\vspace{2pt}
\caption{\textbf{Monitor prompts}, abridged: each query's attached text and
images, prompt, and required JSON reply. Symbols follow \cref{alg:orchestration}.}
\label{fig:monitor_prompts}
\end{figure}

\newcommand{\monimg}[1]{\includegraphics[width=\monimgw]{Figures/monitor_examples/#1}}
\newcommand{\jentry}[1]{\par\hangindent=2.4em\hangafter=1\hspace*{0.6em}#1}
\newcommand{\verdict}[3]{\begin{tcolorbox}[verdictcard, title={#1\hfill#2}]#3\end{tcolorbox}}
\begin{figure}[!t]
\centering
\begin{tabular}{@{}m{\monimgw}@{\hspace{3pt}}m{\monimgw}@{\hspace{6pt}}m{\dimexpr\linewidth-2\monimgw-9pt}@{}}
\centering\cardlabel{Initial} & \centering\cardlabel{Current} & \cardlabel{Monitor verdict (JSON reply)} \\[1pt]
\monimg{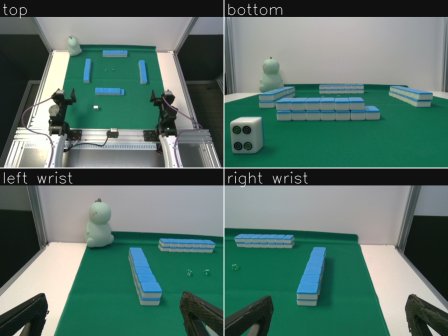} & \monimg{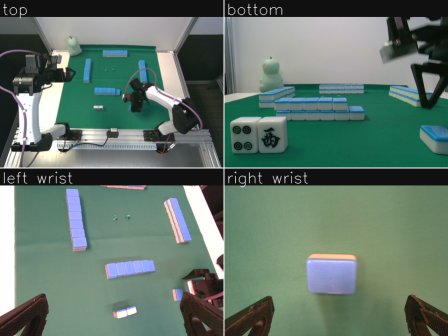} &
\verdict{(a)\enspace Intervention}{$M_{\mathrm{int}}$}{%
\{\jk{"intervene"}: \jl{true},
\jentry{\jk{"scene\_intact"}: \jl{false},}
\jentry{\jk{"reason"}: \js{"The hand row is complete, but an extra tile has been moved loose onto the right side of the table."},}
\jentry{\jk{"recovery"}: \js{"Return the loose extra tile to the tile wall."}\}}} \\[4pt]
\monimg{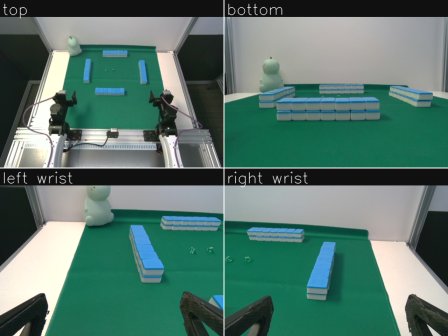} & \monimg{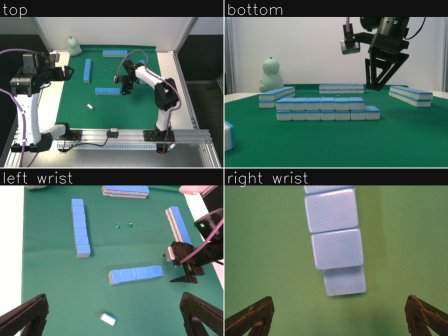} &
\verdict{(b)\enspace Intervention}{$M_{\mathrm{int}}$}{%
\{\jk{"intervene"}: \jl{true},
\jentry{\jk{"scene\_intact"}: \jl{true},}
\jentry{\jk{"reason"}: \js{"The robot has made no progress on re-grasping the deliberately staged tile for over 30 seconds."},}
\jentry{\jk{"recovery"}: \js{""}\}}} \\[4pt]
\monimg{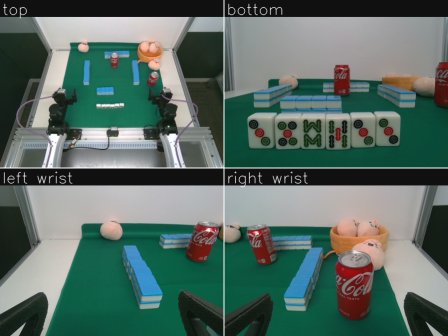} & \monimg{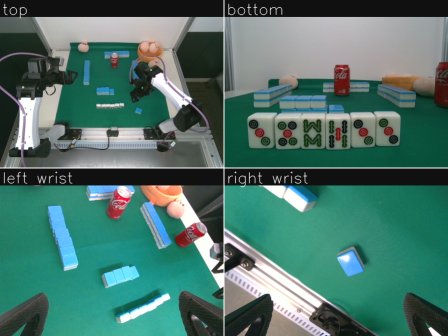} &
\verdict{(c)\enspace Intervention}{$M_{\mathrm{int}}$}{%
\{\jk{"intervene"}: \jl{true},
\jentry{\jk{"scene\_intact"}: \jl{false},}
\jentry{\jk{"reason"}: \js{"The drawn tile has fallen flat after the robot's attempt to stand it upright."},}
\jentry{\jk{"recovery"}: \js{"Stand the fallen tile upright and insert it to the row."}\}}} \\[4pt]
\monimg{c_initial.jpg} & \monimg{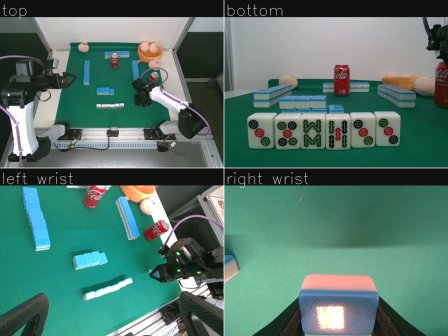} &
\verdict{(c)\enspace Restoration, after recovery}{$M_{\mathrm{res}}$}{%
\{\jk{"restored"}: \jl{true},
\jentry{\jk{"reason"}: \js{"The tiles are upright in the row, and the current arrangement matches the initial workable state."}\}}} \\[4pt]
\monimg{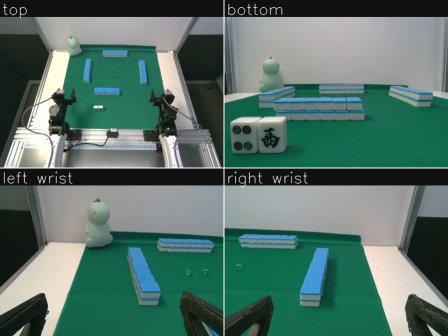} & \monimg{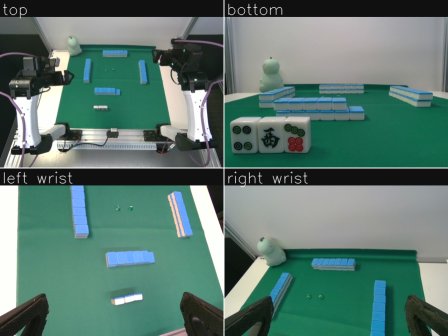} &
\verdict{(d)\enspace Completion}{$M_{\mathrm{done}}$}{%
\{\jk{"complete"}: \jl{true},
\jentry{\jk{"reason"}: \js{"Exactly one upright, forward-facing tile has been added to the right end of the hand, with no fallen tiles and the remaining wall unchanged."}\}}} \\
\end{tabular}
\caption{\textbf{Recorded monitor verdicts} while drawing a mahjong tile. Each
image is the four-camera composite sent to the monitor. (a) The monitor reuses a
known recovery skill verbatim. (b) A stalled task in an intact scene leads to a
task demonstration. (c) The recovery policy executes the proposed instruction,
and the restoration query confirms the scene. (d) The completion query accepts a
finished task.}
\label{fig:monitor_examples}
\end{figure}

\section{Implementation and Evaluation Details}
\label{app:details}

\subsection{Hardware and Tasks}
\label{app:hardware}

Each workstation has two YAM arms, each with six joints and a parallel gripper,
four Intel RealSense cameras recording at $640\times480$ (top, front, and one on
each wrist), and a pair of leader arms through which the operator takes over. A
status display at each station turns green while a policy acts and red when the
operator is needed. \Cref{tab:tasks} lists the four tasks and their success
criteria; each episode is limited to 3 minutes (5,400 control steps). During
parallel collection, one operator supervises four stations, each arranged with
a different background and set of distractor objects (\cref{fig:collection}).
Rollouts start automatically, and drawing sessions continue until the hand
holds eight tiles.

\begin{table}[h]
\centering
\small
\caption{\textbf{Real-robot tasks and success criteria.}}
\label{tab:tasks}
\begin{tabular}{@{}lp{0.72\linewidth}@{}}
\toprule
Task & Success criterion \\
\midrule
Pencil box & The pencil lies in its tray, and the tray is slid back into the box
sleeve. \\
Stack rings & The four rings sit on the peg in size order. \\
Draw tile & Exactly one more tile stands upright at the end of the hand row,
facing the front camera; no tile has fallen, and the tile wall is otherwise
unchanged. \\
Discard tile & The specified hand tile lies face-up in the central discard area,
the other hand tiles are undisturbed, and both grippers are empty. \\
\bottomrule
\end{tabular}
\end{table}

\subsection{Real-Robot Monitor and Data Interface}

The real-robot harness runs a 30\,Hz control loop. The monitor receives all
four camera views (top, front, and both wrists), and the task policy uses the
top and wrist cameras. The task description, completion requirement, and
behavior notes are cached across collection sessions; when no completion
requirement exists yet, the monitor generates it with the robot stationary
before recording starts.

Completion and intervention checks run every 600 and 300 control steps
(20\,s and 10\,s at 30\,Hz), and each autonomous recovery attempt is limited
to 1,500 steps (50\,s). The robot holds still during completion and
restoration checks, while intervention checks run asynchronously. The
intervention query uses the initial scene, strided recent frames, recent
verdicts, and the known recovery instructions, and distinguishes a stalled
task in an intact scene from a scene that needs correction. The restoration
query accepts any arrangement equivalent to the initial one rather than
requiring exact object poses. Before a segment is saved, the operator can
relabel it, which keeps task and recovery data distinct.

A learned-skill registry determines which recovery instructions the recovery
policy handles; after training, newly learned instructions are added to it,
and the online harness reads it. A registered skill receives one bounded
attempt; otherwise, or if restoration fails, the operator takes control.
After two consecutive interventions on a task, the harness requests a new
task demonstration; the count resets after a successful task or a task
demonstration. Successful autonomous recoveries need no human demonstration.

\subsection{Policy Training}
\label{app:training}

All task and recovery policies are fine-tuned from the $\pi_{0.5}$ base
checkpoint \citep{pi2025pi05} with the same recipe (\cref{tab:training}).
Actions are 14-dimensional joint-position targets for both arms, six joints and
one gripper each, predicted in chunks of 50 steps (1.67\,s at 30\,Hz); joint
targets are learned as offsets from the current state, and gripper commands as
absolute values. The policies observe the robot state together with the top and
two wrist images at $224\times224$, and the language prompt is the task
instruction for the task policy and the recovery instruction for the recovery
policy. Training proceeds in stages, each starting from the previous checkpoint
as more data arrive.

\begin{table}[h]
\centering
\small
\caption{\textbf{Policy fine-tuning hyperparameters}, shared by all task and
recovery policies.}
\label{tab:training}
\begin{tabular}{@{}ll@{}}
\toprule
Setting & Value \\
\midrule
Initialization & $\pi_{0.5}$ base checkpoint \\
Actions & 14-D joint positions (joint offsets, absolute grippers) \\
Action chunk & 50 steps (1.67\,s at 30\,Hz) \\
Image inputs & top, left wrist, right wrist; $224\times224$ \\
Optimizer & AdamW, gradient clipping 1.0 \\
Learning rate & cosine decay from $2.5\times10^{-5}$ to $2.5\times10^{-6}$ \\
Warmup & 300--1,000 steps \\
Batch size & 64 \\
EMA decay & 0.99 \\
Steps per stage & 2,000--5,000 \\
Compute & 8 NVIDIA L40 GPUs, 1--3\,h per stage \\
\bottomrule
\end{tabular}
\end{table}

\subsection{Evaluation Protocol}
\label{app:protocol}

\Cref{tab:real} reports 20 trials for each task and configuration. An operator
judges each trial against the task's success criterion in \cref{tab:tasks}.

\subsection{Trajectory Provenance}

Each saved trajectory is stored with its instruction and type: a
\texttt{Recover} trajectory is labeled with its recovery instruction, and
\texttt{Success} and \texttt{Expand} trajectories with the original task. The
session log records the acting controller, episode boundaries, and outcomes,
together with the monitor's images, prompts, responses, and timestamps. Failed
policy prefixes are excluded from the demonstrations but kept in this log, so
every accepted recovery can be audited without using failed segments as
supervision.

\subsection{Qualitative Video Examples}

\Cref{fig:recovery_skills} takes two frames from each of ten recordings: two
real task-policy rollouts, four real recovery-policy rollouts, and four
simulated recovery skills rendered from the digital twin's inspection camera.
Frames use a fixed crop and uniform resizing, without retouching.

\section{Simulated Recovery Skills}
\label{app:skill_gallery}

\begin{figure}[H]
    \centering
    \includegraphics[width=\linewidth]{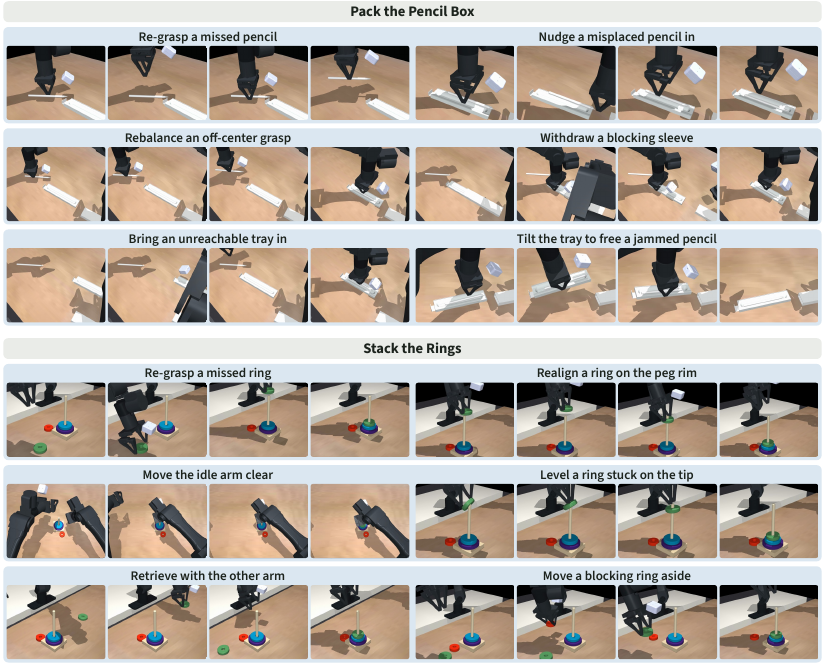}
    \vspace{\capspacebefore}
    \caption{\textbf{Simulated recovery skills for pencil-box packing and
    ring stacking.} Each blue panel shows one recovery skill developed in the
    task's digital twin, as four frames of one rollout, from the failure to
    the result. Frames come from the twin's inspection camera, except
    \emph{Move the idle arm clear} (top camera).}
    \label{fig:skill_gallery_a}
    \vspace{\capspaceafter}
\end{figure}

\begin{figure}[H]
    \centering
    \includegraphics[width=\linewidth]{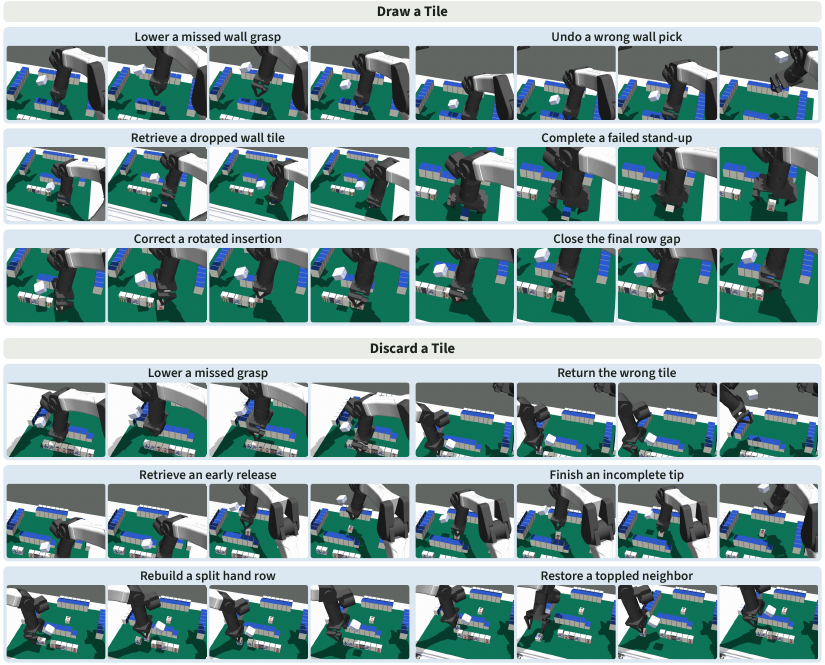}
    \vspace{\capspacebefore}
    \caption{\textbf{Simulated recovery skills for drawing and discarding
    a mahjong tile.} Panels follow the conventions of
    \cref{fig:skill_gallery_a}.}
    \label{fig:skill_gallery_b}
    \vspace{\capspaceafter}
\end{figure}

\Cref{fig:skill_gallery_a,fig:skill_gallery_b} show 24 recovery
skills that the agent developed in the digital twins of the four real-robot
tasks, including the simulated skills pictured in \cref{fig:recovery_skills}.
Each skill appears as four frames of one simulated rollout, from the failure
to the result, taken from the twin's inspection camera; for moving the idle
arm clear, the top camera shows the motion.

\section{Recovery Examples on the Simulation Benchmarks}
\label{app:sim_recovery_gallery}

\begin{figure}[p]
    \centering
    \includegraphics[width=\linewidth,height=0.92\textheight,keepaspectratio]{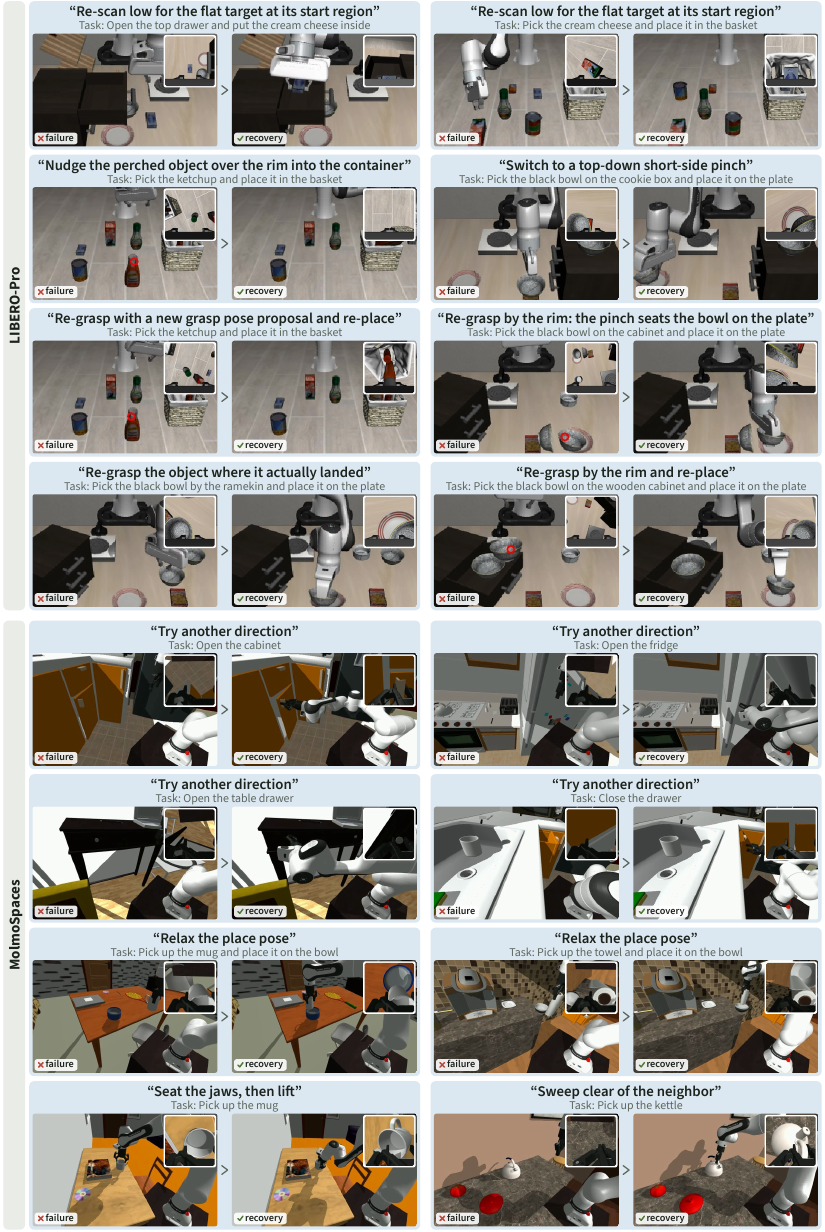}
    \vspace{\capspacebefore}
    \caption{\textbf{Recovery examples on LIBERO-Pro (top) and MolmoSpaces
    (bottom).} Each panel shows one episode's failure and recovery moments
    under its recovery instruction and task, as in \cref{fig:sim_recovery}.
    Red rings in some LIBERO-Pro frames are markers from the source
    recordings.}
    \label{fig:sim_recovery_gallery}
    \vspace{\capspaceafter}
\end{figure}

\Cref{fig:sim_recovery_gallery} extends \cref{fig:sim_recovery} with 16
further episodes, eight from each simulation benchmark. Each episode fails
with recovery disabled and succeeds, from the same initial state, with
recovery enabled. Both frames of an episode come from the run with recovery:
the failure frame just after the setback that triggers the recovery, and the
recovery frame just before the simulator first reports success. The eight
LIBERO-Pro episodes show seven distinct recoveries episodes, and the eight
MolmoSpaces episodes show four recovery skills credited, including ``try another direction''
on four articulated objects. Episodes were chosen
to cover distinct recoveries and scenes. Frames are cropped from the
recordings' camera views and resized uniformly, without retouching.

\end{document}